\documentclass[11pt,a4paper]{article}
\usepackage{EMNLP2023}
\usepackage{times}
\usepackage{latexsym}
\usepackage[T1]{fontenc}
\usepackage[utf8]{inputenc}

\usepackage{multirow}
\usepackage{multicol}
\usepackage{booktabs}
\usepackage{graphicx}
\usepackage{tikz}
\usepackage{paralist}
\usepackage{booktabs}
\usepackage{amsfonts}
\usepackage{CJKutf8}
\usepackage{subfigure}

\usepackage[ruled,vlined]{algorithm2e}
\usepackage{amssymb}
\usepackage{enumitem}
\usepackage{setspace}
\usepackage{amsmath} 
\usepackage{amssymb}
\usepackage{xcolor}

\author{\textbf{
Yanrui Du\textsuperscript{1},
Sendong Zhao\textsuperscript{1}, 
Yuhan Chen\textsuperscript{1}, 
Rai Bai\textsuperscript{1}
} \\
\textbf{
Jing Liu\textsuperscript{2},
Hua Wu\textsuperscript{2} ,
Haifeng Wang\textsuperscript{2},
Bing Qin \textsuperscript{1}
}\\
    \textsuperscript{1}Harbin Institute of Technology, Harbin, China \\  
    \textsuperscript{2}Baidu Inc., Beijing, China\\
    \{ yrdu, sdzhao, yhchen, rbai, bqin\}@ir.hit.edu.cn\\
    \{liujing46, wu\_hua, wanghaifeng\}@baidu.com
}

\title{GLS-CSC: A Simple but Effective Strategy to Mitigate Chinese STM Models' Over-Reliance on Superficial Clue}

\begin{document}
\maketitle
\begin{abstract}

Pre-trained models have achieved success in Chinese Short Text Matching (STM) tasks, but they often rely on superficial clues, leading to a lack of robust predictions. To address this issue, it is crucial to analyze and mitigate the influence of superficial clues on STM models. Our study aims to investigate their over-reliance on the edit distance feature, commonly used to measure the semantic similarity of Chinese text pairs, which can be considered a superficial clue. To mitigate STM models' over-reliance on superficial clues, we propose a novel resampling training strategy called \textbf{\underline{G}radually \underline{L}earn \underline{S}amples \underline{C}ontaining \underline{S}uperficial \underline{C}lue}(\textbf{GLS-CSC}). Through comprehensive evaluations of In-Domain (I.D.), Robustness (Rob.), and Out-Of-Domain (O.O.D.) test sets, we demonstrate that GLS-CSC outperforms existing methods in terms of enhancing the robustness and generalization of Chinese STM models. Moreover, we conduct a detailed analysis of existing methods and reveal their commonality.






\end{abstract}

\begin{CJK*}{UTF8}{gbsn}
\section{Introduction}\label{intro}


The Short Text Matching (STM) task holds significant importance in Natural Language Processing (NLP), aiming to determine the semantic similarity of sentence pairs. The STM task finds widespread applications in areas such as information retrieval, question-answering, and dialogue systems. In recent years, the advent of pre-trained models such as BERT~\cite{devlin2018bert}, ERNIE~\cite{sun2019ernie}, and RoBERTa~\cite{liu2019roberta} established the state-of-the-art models. However, previous studies~\cite{zhang2019paws, mccoy2019right, naik2018stress, nie2020adversarial, huang2020counterfactually, morris2020textattack, zhang2020adversarial, jia2017adversarial} revealed that pre-trained models often rely on superficial clues, resulting in their limited performance on Robustness (Rob.) and Out-of-domain (O.O.D.) test sets.

\begin{figure}[t]
\centering
\includegraphics[scale=0.35]{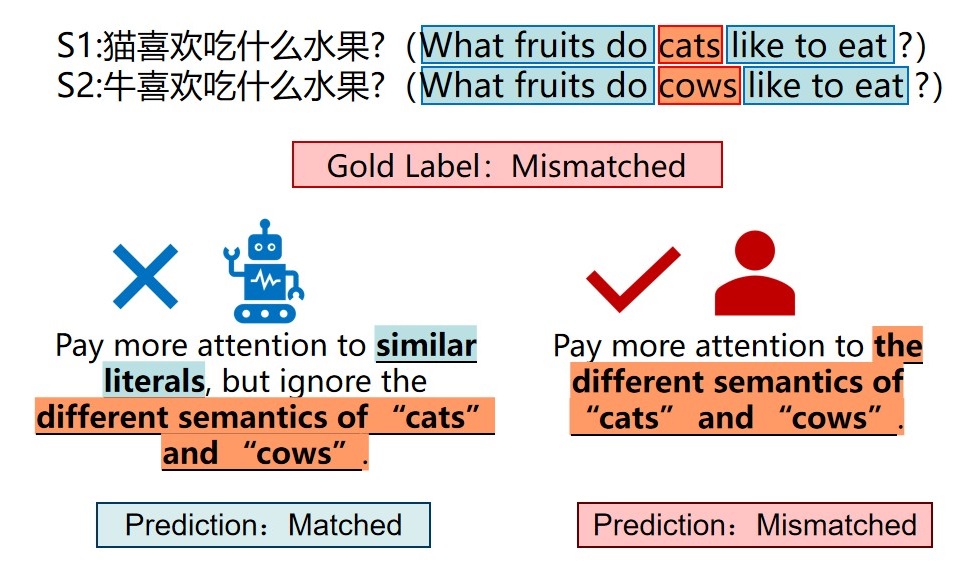}

\caption{An instance of STM models over-relying on  similar textual literals for prediction.}
\label{fig:example}
\end{figure}

Recent studies~\cite{schuster-etal-2019-towards} have highlighted that pre-trained models can easily capture strong correlations as superficial clues due to the biased distribution of the training set. While pre-trained models may achieve satisfactory performance on In-Domain (I.D.) data by leveraging these spurious clues, they remain vulnerable to adversarial samples that exploit superficial clues~\cite{lai2021machine,poliak2018hypothesis,kavumba2019choosing,wang2021identifying}. Fig.\ref{fig:example} illustrates an instance where the STM model predicts semantic matched by focusing on similar textual literals (highlighted in blue) while disregarding semantic differences between ``cat'' and ``cows'' (highlighted in red). Therefore, it becomes imperative to analyze and mitigate the impact of superficial clues on STM models.

For English STM tasks, the concept of word overlap, as a superficial cue that can be easily captured by STM models~\cite{gururangan2018annotation,du2022less}, has garnered significant attention. Word overlap simply examines the presence or absence of words, without taking into account the influence of word order. However, for the Chinese language, extensive linguistics research~\cite {james1985temporal,lapolla1995pragmatic,zhao2016contrastive} consistently demonstrates its sequential nature, where word order strongly influences sentence meaning. For instance, in Chinese, although "喜欢" (like) and "欢喜" (joy) words completely overlap, they have completely different semantics and parts of speech. Compared with word overlap, the edit distance can further consider the influence of word order. Moreover, due to the inherent characteristics of STM tasks, that the divergence in sentence lengths is not excessively significant, it mitigates the drawback of edit distance's sensitivity to disparities in sentence length. To sum up, for Chinese STM tasks, we employ the Levenshtein distance \footnote{https://en.wikipedia.org/wiki/Levenshtein\_distance} as a proxy for the superficial clue to assess STM models' behavior.


In our work, we quantitatively demonstrate that STM models tend to predict semantic matches when the edit distance is relatively small and that STM models tend to predict semantic mismatches when the edit distance is relatively large. Such phenomena suggest that edit distance can serve as a proxy for the superficial clue captured by STM models. For mitigating STM models' over-reliance on the superficial clue, we propose a simple and effective \textbf{GLS-CSC} strategy: \textbf{\underline{G}radually \underline{L}earn \underline{S}amples \underline{C}ontaining \underline{S}uperficial \underline{C}lue}. Similar to curriculum learning~\cite{bengio2009curriculum}, GLS-CSC involves resampling the order of training samples. Furthermore, we conduct our experiments on two Chinese STM datasets (LCQMC$_{train}$ and CCKS$_{train}$). Following the advice of previous work~\cite{geirhos2020shortcut}, we apply multiple O.O.D. real-world test sets to evaluate STM models' performance.  Our experimental results show that our GLS-CSC strategy can mitigate STM models' over-reliance on the superficial clue effectively, and improves their robustness and generalization significantly compared to strong baselines. Notably, for our GLS-CSC strategy, employing edit distance as a proxy has demonstrated notable advantages over employing word overlap. Additionally, we conduct a detailed analysis of existing methods and observe that while they can enhance performance on adversarial data, they inadvertently compromise performance on a specific subset of data.

\section{Related Work}\label{related_work}

\paragraph{Superficial clue.}  Recent studies~\cite{du2022less,schuster-etal-2019-towards} highlight an intriguing observation regarding the tendency of models to rely on superficial clues. Specifically, in NLI (Natural Language Inference) tasks~\cite{gururangan2018annotation}, models often predict the "contradiction" label based on the presence of negative words within text pairs. Similarly, in NLU (Natural Language Understanding) tasks~\cite{lai2021machine}, models have a tendency to extract answers by leveraging the question word type (e.g., utilizing the question word "where" to identify location entities).



\paragraph{Existing methods.} Existing methods can be grouped into three categories: 1. Improve the quality of data. 2. Reweight losses of samples. 3. Debias the model prediction.

\emph{Improve the quality of data.} Several studies~\cite{reddy2019coqa,choi-etal-2018-quac,agrawal-etal-2016-analyzing,zhang-etal-2019-paws,Kaushik2020Learning,mccoy2019right} have implemented various strategies to reduce annotation artifacts in data by imposing guidelines for annotators, thus minimizing the presence of superficial clues. Additionally, some studies~\cite{zellers-etal-2019-hellaswag,sakaguchi2021winogrande,le2020adversarial} filtered samples by adversarial methods to obtain high-quality data as training data. And another study increased the robustness of models by recording forgotten samples during models' training and utilizing them to do secondary training (\textbf{Forg.}). These efforts aim to acquire a set of high-quality training data, although the process can be resource-intensive and may not be universally applicable to all datasets.



\emph{Reweight losses of samples.} Recent studies~\cite{du2022less,schuster-etal-2019-towards} have proposed methods that evaluate samples using statistical information, imposing penalties on those exhibiting strong correlations with labels. Some other studies~\cite{clark2019don,utama2020towards} focused on obtaining a bias-only model, which is then used to score samples and penalize training losses.
 




\emph{Debias the model prediction.} Several studies have explored the integration of output probabilities from a bias-only model into the main model as a means to shift the focus from superficial clues to the true semantics of the text~\cite{clark2019don,he-etal-2019-unlearn,mahabadi2019end}. Notably, two prominent methods in this category are the Product-of-Experts (\textbf{POE}) and Learned-Mixin (\textbf{LEAM}), with the latter being a variant of the former.

\section{Analysis of Superficial Clue}\label{analysis}


Our analysis focuses on the LCQMC dataset~\cite{liu2018lcqmc}, which serves as a large-scale Chinese Short Text Matching dataset. LCQMC comprises two classes: label 1 indicates a semantic match, while label 0 indicates a semantic mismatch. To facilitate our work, we begin by presenting several definitions that will enhance our understanding. Subsequently, we perform two quantitative analyses to investigate the Chinese STM models' over-reliance on the edit distance feature for prediction, leading to the following behaviors.
\begin{itemize}[leftmargin=*,noitemsep,topsep=0pt]
    \item STM models tend to predict label 1 more if the edit distance of text pairs is relatively smaller.
    \item STM models tend to predict label 0 more if the edit distance of text pairs is relatively larger.
\end{itemize}

\begin{figure}[t]
\centering
\includegraphics[scale=0.28]{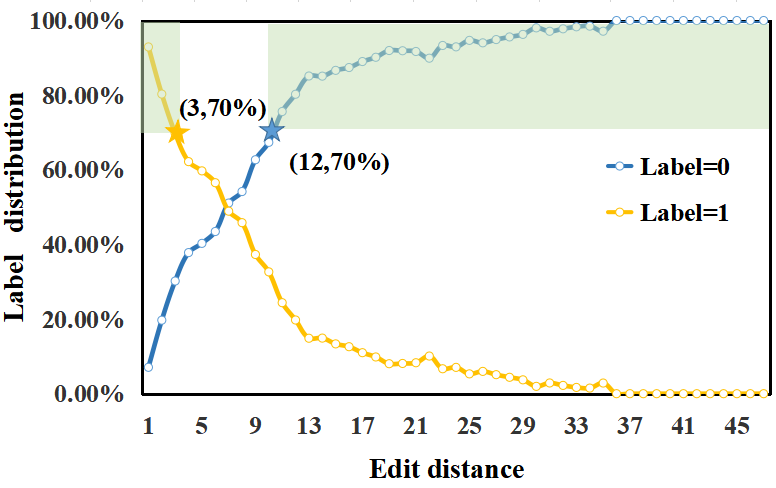}
\caption{For LCQMC$_{train}$, the label distribution of samples at each edit distance.}
\label{fig:train_infor}
\end{figure}

\subsection{Definition.} 
To facilitate our subsequent analysis, we define samples with the strong correlation between edit distance and label as \textbf{Samples \underline{C}ontaining \underline{S}uperficial \underline{C}lue (CSC-Samples)}. As depicted in Fig.~\ref{fig:train_infor}, 70\% is chosen as the threshold for measuring strong correlation and samples covered by green shadow are referred to as CSC-Samples. Moreover, following prior research~\cite{kavumba2019choosing,utama2020towards}, we differentiate between \textbf{easy-to-predict samples} and \textbf{hard-to-predict samples}. Easy-to-predict samples are those where the superficial clue has a positive impact on the models' correct prediction, while hard-to-predict samples are those where the superficial clue has a negative impact on the models' correct prediction. As shown in Tab.\ref{example} (see App.~\ref{app_easy_hard_sample}), easy-to-predict samples include:
\begin{itemize}[leftmargin=*,noitemsep,topsep=0pt]
    \item Samples at relatively smaller edit distance and labeled as 1.
    \item Samples at relatively larger edit distance and labeled as 0.
\end{itemize}
And hard-to-predict samples include: 
\begin{itemize}[leftmargin=*,noitemsep,topsep=0pt]
    \item Samples at relatively smaller edit distance and labeled as 0.
    \item Samples at relatively larger edit distance and labeled as 1.
\end{itemize}

\subsection{Quantitative Analyses}\label{sec_analy}
We verify that Chinese STM models over-rely on the edit distance feature for prediction by conducting the following experiments:
\begin{itemize}[leftmargin=*,noitemsep,topsep=0pt]
    \item Observe the tendency of STM models' prediction as the edit distance of samples changes.
    \item Observe the gap in STM models' performance between easy-to-predict and hard-to-predict samples.
\end{itemize}

\begin{figure}[t]
\centering
\subfigure[Probability that STM models' prediction as 1 on samples at relatively smaller edit distance.]{
\includegraphics[scale=0.45]{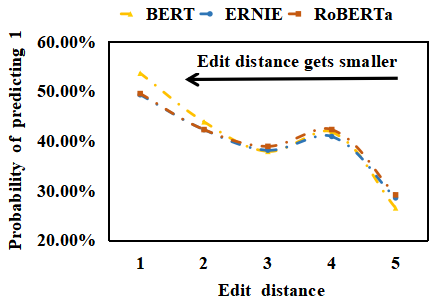}
\label{fig:tend_1}
}
\subfigure[Probability that STM models' prediction as 0 on samples at relatively larger edit distance.]{
\includegraphics[scale=0.45]{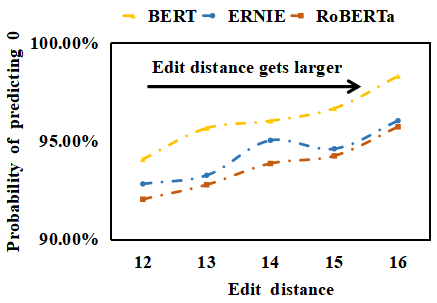}
\label{fig:tend_0}
}
\caption{Tendency of STM models' prediction as the edit distance of sample changes.}
\label{fig:tend_pred}
\end{figure}


\paragraph{Tendency of STM models' prediction.} To analyze the tendency of STM models' prediction, we gather samples with edit distances equal to 1, 2, 3, 4, and 5, respectively, and observe the probability of STM models predicting label 1. Additionally, we collect samples with edit distances equal to 12, 13, 14, 15, and 16, respectively, and observe the probability of STM models predicting label 0. The outcomes, as depicted in Fig.~\ref{fig:tend_1} and Fig.~\ref{fig:tend_0}, corroborate our initial conjecture. Specifically, as the edit distance of the samples decreases, STM models exhibit a tendency to predict label 1. Conversely, as the edit distance of the samples increases, STM models exhibit a tendency to predict label 0.



\paragraph{Gap of STM models' performance.} \label{gap_performance}

To investigate the performance of STM models, we collect easy-to-predict samples as the E-pred set and hard-to-predict samples as the H-pred set. For detailed information on sample collection, please refer to App.~\ref{app_easy_hard_sample}. We train BERT, ERNIE, and RoBERTa models on the LCQMC$_{train}$ dataset and evaluate their performance on the E-pred and H-pred sets. As illustrated in Tab.~\ref{tab:gap_perform}, we observe a substantial performance disparity between the two sets. The STM models exhibit significantly better performance on the E-pred set compared to the H-pred set, with a gap of up to 40\%. This stark contrast aligns with our initial hypothesis: the superficial clue has a positive impact on correct predictions in easy-to-predict samples, while it has a negative impact on on correct predictions in hard-to-predict samples. Notably, the performance of STM models on the H-pred set approaches a random baseline (50\% for the two-classification task), indicating that their over-reliance on the superficial clue compromises their robustness and generalization.

Furthermore, to delve further into the origin of the superficial clue, we exclusively employ CSC-Samples from the LCQMC$_{train}$ dataset to train the STM models. As shown in Tab.~\ref{tab:gap_perform}, the performance gap between the E-pred and H-pred sets widens significantly. The STM models achieve near-perfect performance on the E-pred set, while their performance on the H-pred set plummets close to 0\%. This outcome suggests that the superficial clue is predominantly captured by STM models from CSC-Samples within the LCQMC$_{train}$ dataset.

\begin{table}[t]
\centering
\resizebox{\linewidth}{!}
{
\begin{tabular}{clcc|c}
\toprule[0.7pt]
& Model   & E-pred  & H-pred & $\Delta$ \\
\midrule[0.5pt]
\multirow{3}{*}{All Samples} & BERT    & 90.04\% & 47.27\% & 42.80\% \\
& ERNIE   & 90.42\% & 51.46\% &38.96\% \\
& RoBERTa & 91.46\% & 52.11\% & 39.34\% \\
\midrule[0.5pt]
\multirow{3}{*}{CSC-Samples}& BERT  & 99.45\% & 0.61\% & 98.84\%\\
& ERNIE   & 99.63\% & 0.33\% & 99.30\%\\
& RoBERTa & 99.48\% & 0.81\% & 98.67\%\\
\bottomrule[0.7pt]
\end{tabular}
}
\caption{Train models on all samples and CSC-Samples respectively. Observe the gap in STM models' accuracy between the E-pred set and H-pred set.}
\label{tab:gap_perform}
\end{table}

\section{GLS-CSC Strategy}\label{method}


\begin{table}
\centering
\small
{
\begin{tabular}{c|cc|c}
\toprule[0.7pt]
        & Finetune & GLS-CSC & $\Delta$ \\
\midrule[0.5pt]
BERT    & 86.89\%   & 84.46\%   & -2.43\%                 \\
ERNIE   & 87.46\%     & 84.02\%    & -3.46\%                  \\
RoBERTa & 87.56\%     & 84.82\%    & -2.74\%                 \\
\bottomrule[0.7pt]
\end{tabular}
}
\caption{With LLS-CSC method, STM models' performance on I.D. test sets}
\label{fig:last2learn}
\end{table}

\begin{figure}[t]
\centering
\subfigure{
\includegraphics[scale=0.32]{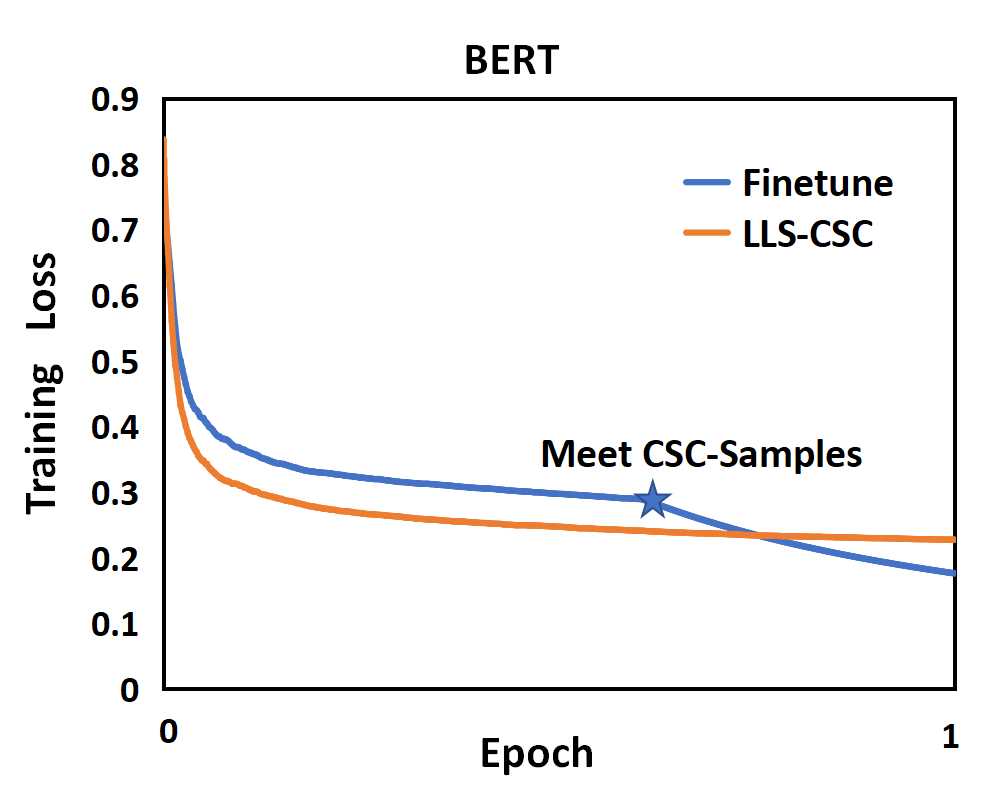}
\label{fig:bert_loss}
}
\caption{Curve of training loss for BERT. The marker point represents that curve drops rapidly after STM models meet CSC-Samples. And curves of training loss for ERNIE and RoBERTa are shown in App.~\ref{app_loss}, we observe a consistent phenomenon.}
\label{fig:training_loss}
\end{figure}

Recent research~\cite{lai2021machine} has highlighted a concerning tendency of pre-trained models to rely heavily on superficial clues early in their learning process, which hinders their ability to explore and develop complex reasoning skills. Interestingly, this phenomenon mirrors a cognitive bias~\footnote{https://en.wikipedia.org/wiki/Cognitive\_bias} observed in humans, who often rely on existing biases instead of actively seeking out further exploration when encountering new information. Following this understanding, we argue that preventing models from capturing superficial clues early on will benefit their subsequent development of advanced reasoning abilities. Therefore, based on the observation that models tend to capture superficial clues from CSC-Samples, we encourage models to last to learn CSC-Samples (called LLS-CSC), placing CSC-Samples at the final stage of the training process. However, the results depicted in Fig.~\ref{fig:last2learn} demonstrate that LLS-CSC will significantly weaken models' performance on the I.D. data (average drop of 2.88 points). 

Moreover, we analyze the reason for its failure. The training loss curves of the models are shown in Fig.~\ref{fig:training_loss} and Fig.~\ref{fig:training_loss_2} (refer to App.~\ref{app_loss}). An obvious phenomenon emerges: when the CSC-Samples are clustered towards the end of the training samples, the loss curve experiences a rapid drop. This observation reveals that STM models still tend to capture superficial clues if the CSC-Samples are clustered locally. Such behavior is detrimental to the learning process of STM models, as it indicates a reliance on less meaningful information. Therefore, for designing a reasonable resampling method, it is crucial to adhere to the following conditions.
\begin{itemize}[leftmargin=*,noitemsep,topsep=0pt]
    \item\textbf{Condition 1:} Tend to make STM models learn CSC-Samples last, so as to avoid STM models from capturing superficial clues in the early stage.
    \item\textbf{Condition 2:} Tend to make CSC-Samples relatively scattered, so as to avoid STM models from capturing superficial clues locally.
\end{itemize}

\begin{algorithm}[h]
\small
\caption{GLS-CSC: Gradually Learn Samples Containing Superficial Clue}
{
\underline{\emph{Define:}} \\
    $T:\mbox{Training set}$ \\
    $T_{csc}:\mbox{CSC-Samples from T}$ \\
    $T_{other}:\mbox{Samples except CSC-Samples from T}$ \\
\underline{\emph{Initialize:}} \\
    $T_{final}\leftarrow\emptyset$ \\
    Set $\alpha$ as shown in Eq.~\ref{eqt:alpha} \\
\underline{\emph{Process of sampling:}} \\
\For{$i = 1~to~Size(T)$}{
    \If {$Size(T_{csc})==0$}{
    Insert $T_{other}$ into $T_{final}$;\\
    Break;
    }
    \If {$Size(T_{other})==0$}{
    Insert $T_{csc}$ into $T_{final}$;\\
    Break;
    }  
    $k \leftarrow 100 \times \alpha \times i$;\\
    $Num \leftarrow RandInt(0,100)$;\\
\If{$Num \ge k $}{
Randomly sample from $T_{other}$;\\
Insert sample into $T_{final}$;
}
\Else{
Randomly sample from $T_{csc}$;\\
Insert sample to $T_{final}$;
}
}
\textbf{return} $T_{final}$
}
\label{algo}
\end{algorithm}


To meet the above conditions, we propose our \textbf{GLS-CSC} strategy. The GLS-CSC strategy effectively controls the proportion of CSC-Samples, that grows linearly during the training process, as shown in Fig.~\ref{fig:proportion_e} (refer to App.~\ref{app_proportion}). The detailed algorithm is outlined in Alg.\ref{algo}. In our algorithm, we divide the training set $T$ into two sets: $T_{csc}$ set and $T_{other}$ set. We need to initialize the parameter $\alpha$, which controls the probability of sampling from $T_{csc}$ set and $T_{other}$ set at each step. Furthermore, we need to ensure that until the final stage of resampling, the remaining samples in $T_{csc}$ or $T_{other}$ are as few as possible. Through our derivation (as shown in App.~\ref{app_derivation}), $\alpha$ will be initialized as follows: 
\begin{gather}\label{eqt:alpha}
\small
\alpha=\frac{2}{(Size(T_{other})/Size(T_{csc})+1) \times Size(T)}
\end{gather}
where $Size$ represents the number of the set.

\section{Experiment}\label{experiment}

\begin{table*}[ht]
\centering
\resizebox{\linewidth}{!}
{
\begin{tabular}{clcc|cc|ccccc}
\toprule[0.7pt]
\multirow{2}{*}{LCQMC$_{train}$} &            & \multicolumn{2}{c}{\textbf{I.D.}}                                     & \multicolumn{2}{c}{\textbf{Rob.}}                                    & \multicolumn{5}{c}{\textbf{O.O.D.}}                                                                                                                            \\

                          &          & $\underline{\textbf{LCQMC}}$ & $\Delta$ & $\textbf{DuQM}$ & $\Delta$ & $\underline{\textbf{OPPO}}$ &  $\underline{\textbf{BQ}}$ & $\underline{\textbf{CHIP-STS}}$ & $\overline{\textbf{AVG}}$ & $\Delta$ \\
\midrule[0.5pt]
         \multirow{9}{*}{BERT}                 & Finetune$^{\clubsuit}$ & 86.89\%                               & {-} & 67.19\%                              & {-} & 81.07\%                              & 57.93\%                            & 65.84\%                             & 72.93\%      & {-} \\
                          & Forg.$^{\clubsuit}$    & 86.49\%                               & -0.40\%              & 68.44\%                              & 1.25\%               & 80.91\%                              & 58.57\%                            & 65.09\%                             & 72.76\%      & -0.17\%              \\
                          & Rew.$_{self}^{\clubsuit}$     & 87.00\%                               & 0.11\%               & 68.41\%                              & 1.22\%               & 80.90\%                              & 57.97\%                            & 65.48\%                             & 72.84\%      & -0.09\%              \\
                          & POE$_{self}^{\clubsuit}$       & 86.77\%                               & -0.12\%              & 67.52\%                              & 0.33\%              & 80.74\%                              & 57.55\%                            & 66.31\%                             & 72.84\%      & -0.09\%              \\
                          & LEAM$_{self}^{\clubsuit}$      & 86.85\%                               & -0.04\%              & 67.35\%                              & 0.16\%               & 80.65\%                              & 58.06\%                            & 66.69\%                             & 73.06\%      & 0.13\%               \\
                          & CL$^{\clubsuit}$  & 86.99\%                  & 0.10\%              & 68.18\%                              & 0.99\%               & 81.02\%                              & 58.26\%                            & 66.12\%                             & 73.10\%      & 0.17\%              \\
            \cline{2-11}
                          & Rew.$_{our}^{\spadesuit}$     & 87.74\%                               & 0.85\%               & 69.54\%                              & 2.35\%               & 81.08\%                              & 57.72\%                            & 64.55\%                             & 72.77\%      & -0.16\%              \\
                          & POE$_{our}^{\spadesuit}$       & 88.04\%                               & 1.15\%               & 67.42\%                              & 0.23\%               & 80.74\%                              & 54.99\%                            & 62.96\%                             & 71.68\%      & -1.25\%              \\
                          & LEAM$_{our}^{\spadesuit}$      & 86.97\%                               & 0.08\%               & 68.22\%                              & 1.03\%               & 80.76\%                              & 57.12\%                            & 65.71\%                             & 72.64\%      & -0.29\%              \\
            \cline{2-11}
                        & GLS-CSC$^{\dagger}$  & 86.90\% & 0.01\%   & 68.97\%  & 1.78\%  & 81.60\% & 58.80\% & 67.45\% & 73.69\%  & 0.76\%\\
                        
                       & GLS-CSC$^{\ddagger}$    & 87.64\%                               & 0.75\%               & 72.32\%                              & 5.13\%               & 81.99\%                              & 58.70\%                            & 67.79\%                             & 74.03\%      & 1.10\%               \\
\midrule[0.5pt]
            \multirow{2}{*}{ERNIE}  & Finetune$^{\clubsuit}$ & 87.46\%                               & {-} & 71.29\%                              & {-} & 82.40\%                              & 60.05\%                            & 66.81\%                             & 74.18\%      & {-} \\
            & GLS-CSC$^{\dagger}$  & 87.57\% & 0.11\%   &73.37\%  & 2.08\%  & 82.30\% & 60.44\% & 66.44\% & 74.29\%  & 0.11\%\\

                   & GLS-CSC$^{\ddagger}$     & 87.61\%                               & 0.15\%               & 75.00\%                              & 3.71\%               & 82.21\%                              & 60.60\%                            & 67.98\%                             & 74.60\%      & 0.42\%               \\
\midrule[0.5pt]
     \multirow{2}{*}{RoBERTa}     & Finetune$^{\clubsuit}$ & 87.56\%                               & {-} & 73.07\%                              & {-} & 82.61\%                              & 60.08\%                            & 68.68\%                             & 74.73\%      & {-} \\

 & GLS-CSC$^{\dagger}$     & 87.69\%                               & 0.13\%               & 74.35\%                              & 1.28\%               & 82.45\%                              & 61.74\%                            & 69.45\%                             & 75.33\%      & 0.60\%    \\

 & GLS-CSC$^{\ddagger}$     & 87.89\%                               & 0.33\%               & 75.62\%                              & 2.55\%               & 82.68\%                              & 61.23\%                            & 69.14\%                             & 75.24\%      & 0.51\%    \\
\bottomrule[0.7pt]
\end{tabular}
}
\caption{Experiment on LCQMC$_{train}$. Train BERT, ERNIE and RoBERTa with previous methods and our GLS-CSC method, and evaluate STM models' accuracy on LCQMC, DuQM, OPPO, BQ, and CHIP-STS. $\Delta$ represents the gap between other methods and finetune baseline. $\overline{\textbf{AVG}}$ represents the average of models' performance on $\underline{\textbf{LCQMC}}$, $\underline{\textbf{OPPO}}$, $\underline{\textbf{BQ}}$ and $\underline{\textbf{CHIP-STS}}$, for measuring STM models' generalization. $^{\dagger}$ and $^{\ddagger}$ represent the GLS-CSC strategy with word overlap and edit distance as proxies for the superficial clue respectively.}
\label{lcqmc_result}
\end{table*}

\begin{table*}[ht]
\centering
\resizebox{\linewidth}{!}
{
\begin{tabular}{clcc|cc|ccccc}
\toprule[0.7pt]
\multirow{2}{*}{CCKS$_{train}$} &            & \multicolumn{2}{c}{\textbf{I.D.}}                                     & \multicolumn{2}{c}{\textbf{Rob.}}                                    & \multicolumn{5}{c}{\textbf{O.O.D.}}                                                                                                                            \\

                          &          & $\underline{\textbf{CCKS}}$ & $\Delta$ & \textbf{DUQM} & $\Delta$ & $\underline{\textbf{OPPO}}$ &  $\underline{\textbf{BQ}}$ & $\underline{\textbf{CHIP-STS}}$ & $\overline{\textbf{AVG}}$ & $\Delta$    \\
\midrule[0.5pt]
         \multirow{9}{*}{BERT}                 & Finetune$^{\clubsuit}$ & 94.13\%                               & {-} & 29.78\%                              & {-} & 49.69\%                              & 84.71\%                            & 70.70\%                             & 74.81\%      & {-} \\
                          & Forg.$^{\clubsuit}$    & 93.92\%                               & -0.21\%              & 30.45\%                              & 0.67\%               & 50.17\%                              & 84.03\%                            & 69.58\%                             & 74.43\%      & -0.38\%              \\
                          & Rew.$_{self}^{\clubsuit}$     & 93.77\%                               & -0.36\%               & 30.02\%                              & 0.24\%               & 50.70\%                              & 81.67\%                            & 69.85\%                             & 74.00\%      & -0.81\%              \\
                          & POE$_{self}^{\clubsuit}$       & 92.30\%                               & -1.83\%              & 29.53\%                              & -0.25\%              & 49.29\%                              & 83.58\%                            & 69.76\%                             & 73.73\%      & -1.08\%              \\
                          & LEAM$_{self}^{\clubsuit}$      & 93.01\%                               & -1.12\%              & 30.61\%                              & 0.83\%               & 51.06\%                              & 84.01\%                            & 69.43\%                             & 74.13\%      & -0.43\%               \\
                           & CL$^{\clubsuit}$      & 94.00\%                               & -0.13\%              & 30.25\%                              & 0.47\%               & 51.29\%                              & 84.96\%                            & 69.88\%                             & 75.03\%      & 0.22\%               \\
            \cline{2-11}
                          & Rew.$_{our}^{\spadesuit}$     & 94.08\%                               & -0.05\%               & 30.31\%                              & 0.53\%               & 51.35\%                              & 84.52\%                            & 71.08\%                             & 75.26\%      & 0.45\%              \\
                          & POE$_{our}^{\spadesuit}$       & 92.27\%                               & -1.86\%               & 29.35\%                              & -0.43\%               & 48.62\%                              & 84.25\%                            & 70.22\%                             & 73.84\%      & -0.97\%              \\
                          & LEAM$_{our}^{\spadesuit}$      & 93.10\%                               & -1.03\%               & 30.47\%                              & 0.69\%               & 49.24\%                              & 84.04\%                            & 70.49\%                             & 73.72\%      & -0.59\%              \\
            \cline{2-11}
                     & GLS-CSC$^{\dagger}$     & 93.37\%            & -0.76 \%          & 30.47\%               & 0.69\%               & 50.96\%                              & 85.14\%                            & 70.33\%                 &74.95\%          & 0.14\%               \\

                       & GLS-CSC$^{\ddagger}$     & 93.40\%                               & -0.73\%               & 31.26\%                              & 1.48\%               & 52.32\%                              & 84.73\%                            & 70.42\%                             & 75.22\%      & 0.41\%               \\
\midrule[0.5pt]
            \multirow{2}{*}{ERNIE}  & Finetune$^{\clubsuit}$ & 93.15\%                               & {-} & 33.64\%                              & {-} & 52.76\%                              & 85.61\%                            & 73.76\%                             & 76.32\%      & {-} \\

            & GLS-CSC$^{\dagger}$     & 92.19\%            & -0.96\%          & 35.58\%               & 1.94\%               & 56.09\%                              & 84.77\%                            & 73.93\%                 &76.75\%          & 0.43\%               \\
                        
                   & GLS-CSC$^{\ddagger}$    & 92.00\%                               & -1.15\%               & 38.17\%                              & 4.53\%               & 57.37\%                              & 85.08\%                            & 74.00\%                             & 77.11\%      & 0.79\%               \\
\midrule[0.5pt]
     \multirow{2}{*}{RoBERTa}     & Finetune$^{\clubsuit}$ & 93.94\%                               & {-} & 32.48\%                              & {-} & 51.85\%                              & 85.57\%                            & 72.20\%                             & 75.89\%      & {-} \\

     & GLS-CSC$^{\dagger}$     & 92.53\%            & -1.41\%          & 34.27\%               & 1.79\%               & 53.98\%                              & 84.85\%                            & 73.36\%                 &75.93\%          & 0.04\%               \\

 & GLS-CSC$^{\ddagger}$    & 92.84\%                               & -1.10\%               & 35.94\%                         & 3.46\%               & 54.36\%                              & 85.10\%                            & 71.79\%                             & 76.02\%      & 0.13\%    \\
\bottomrule[0.7pt]
\end{tabular}
}
\caption{Experiment on CCKS$_{train}$. Train BERT, ERNIE and RoBERTa with previous methods and our GLS-CSC method, and evaluate STM models' accuracy on CCKS, DuQM, OPPO, BQ, and CHIP-STS. The meaning of symbols is consistent with Tab.~\ref{lcqmc_result}.}
\label{ccks_result}
\end{table*}



\subsection{Preliminary Preparation}\label{pre_prepar}
\paragraph{Datasets.} In our experiments, we train STM models on general domain training sets and evaluate their performance on different domain test sets. Specifically, we regard LCQMC$_{train}$ and CCKS$_{train}$ as training sets. LCMQC~\cite{liu2018lcqmc} is the Chinese largest-scale STM dataset, which is collected from BaiduZhidao and CCKS\footnote{https://www.biendata.xyz/competition/} is a campus STM dataset collected by Sohu. For our evaluation, we observe STM models' robustness on DuQM test set, which is a fine-grained controlled robustness dataset. And, we regard OPPO, BQ, and CHIP-STS as our O.O.D. test sets. OPPO\footnote{https://luge.ai} is collected from OPPO XiaoBu Dialogue application. BQ~\cite{chen-etal-2018-bq} is a financial domain dataset from the service logs of bank. CHIP-STS is a medical domain dataset from CBLUE benchmark~\cite{zhang2021cblue}.  Our strength is that we select multiple real-world datasets from different domains to provide a comprehensive and detailed evaluation. Statistics of datasets used in our experiment are presented in Tab.~\ref{app:statistic}(in Appendix). And in App.~\ref{app_spearman}, we calculate the Spearman's correlation between them, which shows their difference.

\paragraph{Training details.} We use integrated interface BertForSequenceClassification\footnote{https://huggingface.co/docs/transformers.} from huggingface to load pre-trained models, including BERT$_{base}$, ERNIE$_{base}$ and RoBERTa$_{large}$. We set the batch size as 64 and the proportion of weight decay as 0.01. For BERT and ERNIE, the learning rate is 2e-5 and for RoBERTa, the learning rate is 5e-6. We train models for 3 epochs on LCQMC$_{train}$ and for 5 epochs on CCKS$_{train}$. For the process of training, we check the performance of models on the validation set every 500 steps to select the best performance checkpoint as our main model. Additionally, for our experimental results, we report the average of three different seeds and our improvements are statistically significant with a p-value of paired t-test less than 0.05.

\paragraph{Metrics.} We utilize accuracy as the metric to assess the STM models’ performance. And we consider the STM models' performance on the robustness DuQM test set as a measure of their robustness, while the average performance on the I.D. test set and the three O.O.D. test sets serve as a measure of their generalization.


\paragraph{Baseline.} We select Forg., Rew.$_{self}$, POE$_{self}$, and LEAM$_{self}$ as our strong baselines and reimplement them through open-sourced code from previous work. Additionally, as described in Sec.~\ref{sec_analy}, we train our own bias-only model exclusively on CSC-Samples. And as mentioned in Section \ref{related_work}, Forg., Rew.$_{self}$, POE$_{self}$, and LEAM$_{self}$ require the ensemble of a bias-only model. Therefore, we apply our bias-only model to these methods and refer to them as Rew.$_{our}$, POE$_{our}$, and LEAM$_{our}$. We also select the curriculum learning (CL) method as our baseline, which shares a common idea of resampling. The CL method focuses on learning from simple to difficult samples. In our work, we consider short text pairs, which contain less information, as simple samples, and long text pairs, which contain more information, as difficult samples. Furthermore, we explore the use of word overlap as a proxy for the superficial clue, instead of edit distance, to apply our GLS-CSC strategy.

\begin{table*}[ht]
\centering
\resizebox{\linewidth}{!}{
\begin{tabular}{cl|cc|cc|cc|cc|cc|cc}
\toprule[0.7pt]
 &              & \multicolumn{6}{c|}{\textbf{LCQMC$_{train}$}}                                                                                                                                 & \multicolumn{6}{c}{\textbf{CCKS$_{train}$}}                                                                                                                                  \\
                          &          & \multicolumn{1}{c}{\textbf{E-pred}} & \textbf{$\Delta$} & \textbf{H-pred}  & \textbf{$\Delta$} & \textbf{Normal} & \textbf{$\Delta$} & \textbf{E-pred} & \textbf{$\Delta$} & \textbf{H-pred} & \textbf{$\Delta$} & \textbf{Normal} & \textbf{$\Delta$} \\
 \midrule[0.5pt]                        
 \multirow{9}{*}{BERT}     & Finetune$^{\clubsuit}$ & 90.04\%           & {-}          & 47.27\%           & {-}          & 64.46\%           & {-}          & 88.95\%           & {-}          & 27.17\%                 & {-}    & 70.43\%                 & {-}    \\
                          & Forg.$^{\clubsuit}$    & 90.06\%           & 0.02\%                        & 48.41\%           & 1.14\%                        & 64.78\%           & 0.32\%                        & 87.10\%           & -1.85\%                       & 28.77\%                 & 1.60\%                  & 70.15\%                 & -0.28\%                 \\
                          & Rew.$_{self}^{\clubsuit}$     & 89.86\%           & -0.18\%                       & 48.60\%           & 1.33\%                        & 64.24\%           & -0.22\%                       & 88.20\%           & -0.75\%                       & 27.82\%                 & 0.65\%                  & 70.84\%                 & 0.41\%                  \\
                          & POE$_{self}^{\clubsuit}$      & 90.38\%           & 0.34\%                        & 46.12\%           & -1.15\%                       & 63.87\%           & -0.59\%                       & 87.31\%           & -1.64\%                       & 26.96\%                 & -0.21\%                 & 69.80\%                 & -0.63\%                 \\
                          & LEAM$_{self}^{\clubsuit}$     & 89.64\%           & -0.40\%                       & 48.13\%           & 0.86\%                        & 64.32\%           & -0.14\%                       & 87.91\%           & -1.04\%                       & 28.03\%                 & 0.86\%                  & 70.61\%                 & 0.18\%                  \\
                          & CL$^{\clubsuit}$     & 91.02\%           & 0.98\%                       & 47.28\%           & 0.01\%                        & 64.77\%           & 0.31\%                       & 87.63\%           & -1.32\%                       & 28.99\%                 & 1.82\%                  & 71.13\%                 & 0.70\%                  \\
                          \cline{2-14}
                          & Rew.$_{our}^{\spadesuit}$     & 89.20\%           & -0.84\%                       & 49.94\%           & 2.67\%                        & 64.03\%           & -0.43\%                       & 88.59\%           & -0.36\%                       & 27.91\%                 & 0.74\%                  & 71.13\%                 & 0.70\%                  \\
                          & POE$_{our}^{\spadesuit}$      & 87.52\%           & -2.52\%                       & 48.73\%           & 1.46\%                        & 61.47\%           & -2.99\%                       & 86.54\%           & -2.41\%                       & 28.21\%                 & 1.04\%                  & 69.82\%                 & -0.61\%                 \\
                          & LEAM$_{our}^{\spadesuit}$     & 89.61\%           & -0.43\%                       & 48.18\%           & 0.91\%                        & 63.69\%           & -0.77\%                       & 87.54\%           & -1.41\%                       & 28.10\%                 & 0.93\%                  & 70.06\%                 & -0.37\%                 \\
                          \cline{2-14}

      
   & GLS-CSC   & 88.04\%           & -2.00\%                       & 54.57\%           & 7.30\%                        & 65.23\%           & 0.77\%                        & 86.46\%           & -2.49\%                       & 30.61\%                 & 3.44\%                  & 71.65\%                 & 1.22\%                  \\
\midrule[0.5pt]
 \multirow{2}{*}{ERNIE}     & Finetune$^{\clubsuit}$ & 90.42\%           & {-}          & 51.46\%           & {-}          & 66.64\%           & {-}          & 89.32\%           & {-}          & 31.71\%                 & {-}    & 72.35\%                 & {-}    \\
   
   
   & GLS-CSC    & 87.74\%           & -2.68\%                       & 57.55\%           & 6.09\%                        & 66.93\%           & 0.29\%                       & 87.35\%           & -1.97\%                       & 37.66\%                 & 5.95\%                  & 73.81\%                 & 1.46\%                  \\
\midrule[0.5pt]
\multirow{2}{*}{RoBERTa} 
                          & Finetune$^{\clubsuit}$ & 91.46\%           & {-}          & 52.11\%           & {-}          & 67.06\%           & {-}          & 88.54\%           & {-}          & 30.30\%                 & {-}    & 72.26\%                 & {-}    \\


 & GLS-CSC    & 88.30\%           & -3.16\%                       & 58.67\%           & 6.56\%                        & 67.29\%           & 0.23\%                        & 86.36\%           & -2.18\%                       & 35.19\%                 & 4.89\%                  & 72.86\%                 & 0.60\%                  \\
\bottomrule[0.7pt]
\end{tabular}
}
\caption{Train STM models on LCQMC$_{train}$ and CCKS$_{train}$, we evaluate their performance on E-pred, H-pred and Normal test sets. The meaning of symbols is consistent with Tab.~\ref{lcqmc_result}.}
\label{detail_analysis}
\end{table*}

\subsection{Main Experiment}\label{main_exp}

In our experiments, we evaluate the performance of STM models with different methods on BERT. Additionally, we conduct experiments to validate the effectiveness of our GLS-CSC method on other pre-trained models, such as ERNIE and RoBERTa.


\paragraph{Trained on LCQMC$_{train}$.} As shown in Tab.~\ref{lcqmc_result}, for BERT, methods$^{\clubsuit}$ improve STM models' performance by approximately 1\% on the Rob. test set, while remaining their performance on the I.D. test set and generalization ($\overline{AVG}$). Methods$^{\spadesuit}$ which ensemble our bias-only model, result in greater improvements on the I.D. and Rob. test sets, but albeit with a slight decrease in generalization. In comparison, our GLS-CSC$^{\ddagger}$ method achieves more significant improvements in both the robustness and generalization (up to 5.13\% and 1.10\%) and demonstrates notable improvement (0.75\%) on the I.D. test set. Furthermore, it is worth noting that even when using word overlap as a proxy for the superficial clue, our GLS-CSC$^{\dagger}$ method still yields considerable enhancements, although not as prominent as when using edit distance (GLS-CSC$^{\ddagger}$). This observation underscores the notion that edit distance may be a more suitable proxy for the superficial clue in Chinese STM tasks. For ERNIE and RoBERTa, our GLS-CSC ($^{\dagger}$ and $^{\ddagger}$) method consistently achieves significant improvements in both robustness and generalization, while maintaining strong performance on the I.D. test set.

\paragraph{Trained on CCKS$_{train}$.} As shown in Tab.~\ref{ccks_result}, most of the previous methods$^{\clubsuit}$ demonstrate improvements in the robustness of STM models, but at the cost of weakened generalization ($\overline{\textbf{AVG}}$). Moreover, the POE and LEAM methods even exhibit a negative impact, indicating their lack of robustness. In contrast, our GLS-CSC$^{\ddagger}$ method achieves significant improvements in both the robustness and generalization of STM models. For the CCKS dataset, we still observe steady improvements with both GLS-CSC$^{\dagger}$ and GLS-CSC$^{\ddagger}$ methods, although GLS-CSC$^{\dagger}$ shows comparatively weaker performance improvement. However, on the I.D. test set, all methods experience a slight decrease in performance, which may be attributed to inherent issues with the data distribution itself.

 Overall, our strategy is simple and effective, bringing more significant improvements in the robustness and generalization of STM models. Furthermore, our strategy is flexible, achieving varying degrees of performance improvement when facing different proxies for superficial clues. Through our experiments, we reveal that compared to word overlap, edit distance may be a better proxy for the superficial clue in Chinese STM tasks.

\begin{figure}[ht]
\centering
\subfigure{
\includegraphics[scale=0.32]{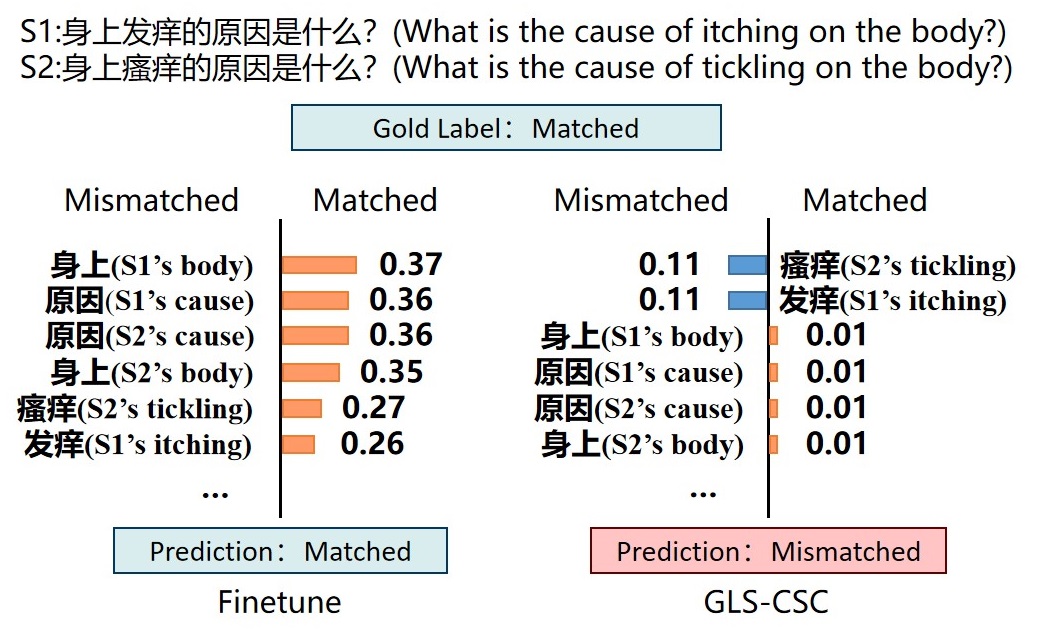}
}
\caption{Explain the model behavior with the LIME method, the figure shows the contribution of words to the predicted label.}
\label{fig:case_study}
\end{figure}

\subsection{Analysis Experiment}\label{analysis_exp}

To gain a deeper understanding of existing methods, we further examine STM models’ performance on the E-pred, H-pred, and Normal test sets (as mentioned in App.~\ref{app_easy_hard_sample}). As shown in Tab.~\ref{detail_analysis}, a common observation across most methods is that while STM models tend to perform better on the H-pred set, this improvement often comes at the expense of their performance on the E-pred set. We think this is a reasonable phenomenon. In easy-to-predict samples, the superficial clue plays a significant role in facilitating the models' correct predictions. Therefore, when attempting to mitigate the models' over-reliance on the superficial clue, it is reasonable to expect a decrease in performance on the E-pred set. With the interpretability algorithm LIME~\cite{ribeiro2016should}, we provide a case study to illustrate this phenomenon in Fig.~\ref{fig:case_study}. Although the original model achieves correct prediction, it tends to place excessive focus on identical words within text pairs, disregarding key semantic information. Conversely, the model with our GLS-CSC strategy, despite achieving incorrect prediction, focuses more on key semantic information. However, a notable limitation lies in the understanding of the semantic relationship between "tickling" and "itching". Overall, our observation quantitatively explores the fact that the superficial clue can lead to a false sense of high performance.

\begin{figure}[t]
\centering
\subfigure{
\includegraphics[scale=0.3]{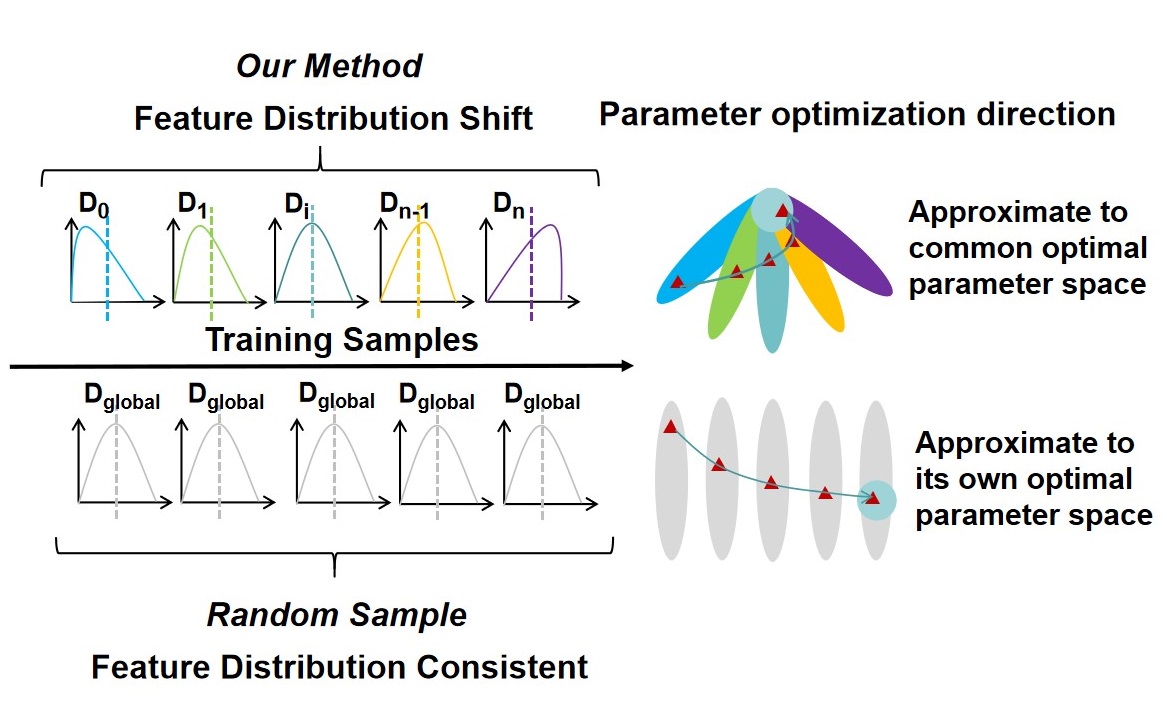}
}
\caption{Explain the effectiveness of our GLS-CSC method from the perspective of feature distribution. The left shows the edit distance feature distribution of local training data and the right shows the optimization direction in parameter space.}
\label{fig:interpre_method}
\end{figure}

\subsection{Another Perspective}\label{perspec}
We provide an explanation for the effectiveness of our GLS-CSC strategy based on the perspective of feature distribution. In random sampling, we assume that the feature distribution of the local training data aligns with the global feature distribution $D_{global}$, represented as $D_{global}$=\{$D_{global}$,...,$D_{global}$\}. However, in our GLS-CSC strategy, we introduce the edit distance feature as a superficial clue and resample the training data order accordingly. This results in a change in the edit distance feature distribution of the local training data, progressing from $D_{0}$ to $D_{n}$ as shown in Fig.\ref{fig:interpre_method}. In random sampling, where the feature distribution remains consistent, the model's parameter optimization approximates the optimal space under $D_{global}$, leading to suboptimal performance under other feature distributions. In contrast, our GLS-CSC method guides the model's parameter optimization to converge towards a \textbf{common} optimal space among the different feature distributions \{$D_{0}$,...,$D_{n}$\}, which improves the model's performance across various feature distributions.

\section{Conclusion}\label{conclu}



In our study, we focus on the issue of STM models' over-reliance on superficial clues in Chinese short text matching tasks. With edit distance as a proxy for superficial clues, we provide quantitative analysis to explore the behavior of STM models. Based on our analysis, we propose a simple and effective strategy called GLS-CSC, which mitigates Chinese STM models' over-reliance on the superficial clue, thereby significantly enhancing their robustness and generalization. Furthermore, we conduct in-depth analyses to gain insights into the behavior of the STM models and provide a comprehensive understanding.

\section{Limitations}

In our work, we analyze the superficial clue for the Chinese STM tasks, and propose our resampling GLS-CSC strategy to alleviate the model's over-reliance on the superficial clue. However, we believe that our work still has the following limitations:
\begin{itemize}[leftmargin=*,noitemsep,topsep=0pt]
\item Our GLS-CSC strategy is only suitable for scenarios where the superficial clue has been explored, while for scenarios where the superficial clue is unknown, how to adapt our method still needs to be further explored.
\item We explore the superficial clue for STM tasks based on our statistical analysis, but how to automatically explore the superficial clue for different tasks is still a major problem.
\end{itemize}

\section{Ethics Statement}
In our work, we conduct all experiments on publicly available datasets for STM tasks with authorization from the respective maintainers.



\bibliographystyle{acl_natbib}
\bibliography{custom}

\clearpage
\appendix

\section{Easy-to-capture Samples and Hard-to-predict Samples.}\label{app_easy_hard_sample}
In Tab.~\ref{example}, we give examples of easy-to-capture samples and hard-to-predict samples. And we describe in detail how we collect easy-to-predict samples and hard-to-predict samples. As we can observe in Fig.~\ref{analysis}, edit distance $\leq$ 3 and $\geq$ 12 are regarded as the boundary for CSC-Samples. For easy-to-predict samples and hard-to-predict samples, we choose the same boundary. From DuQM, OPPO, BQ, and CHIP-STS test sets, we collect samples with edit distance $\leq$ 3 and labeled as 1, and samples with edit distance $\geq$ 12 and labeled as 0, which are regarded as the E-pred set. And we collect samples with edit distance $\leq$ 3 and labeled as 0, and samples with edit distance $\geq$ 12 and labeled as 1, which are regarded as the H-pred set. The remaining samples are regarded as the Normal test set. The E-pred set contains 11,008 samples, the H-pred set contains 12,391 samples and the Normal set contains 20,722 samples.


\section{Training Loss.}\label{app_loss}
\begin{figure}[t]
\centering
\subfigure{
\includegraphics[scale=0.37]{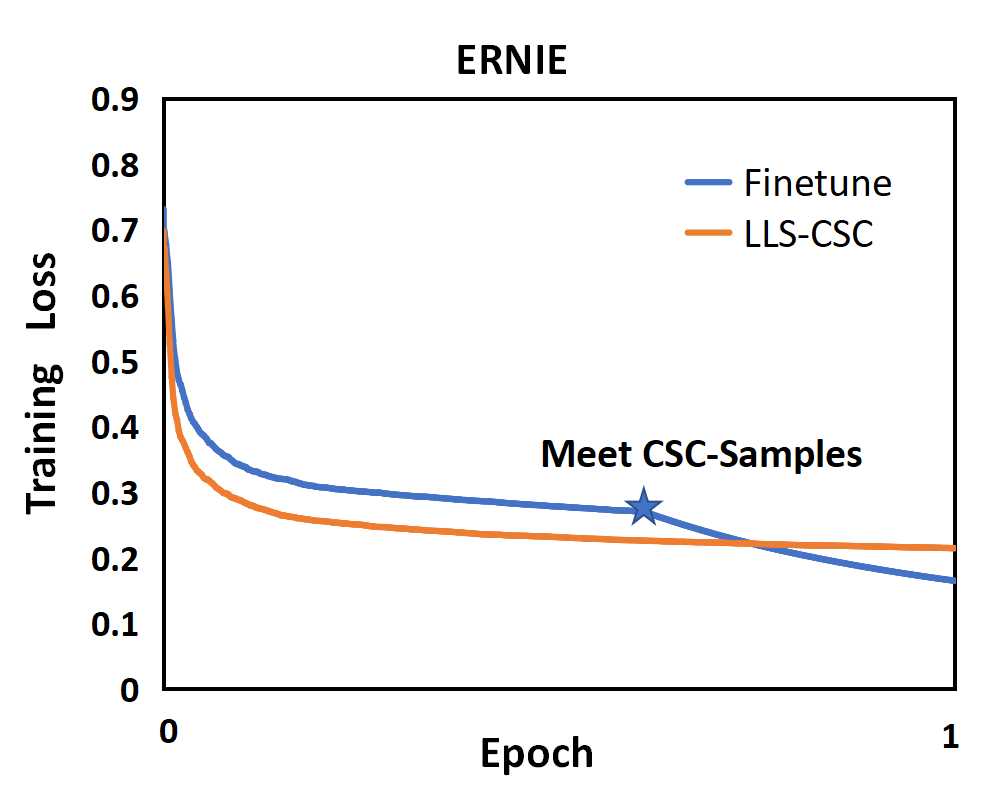}
\label{fig:ernie_loss}
}
\subfigure{
\includegraphics[scale=0.3]{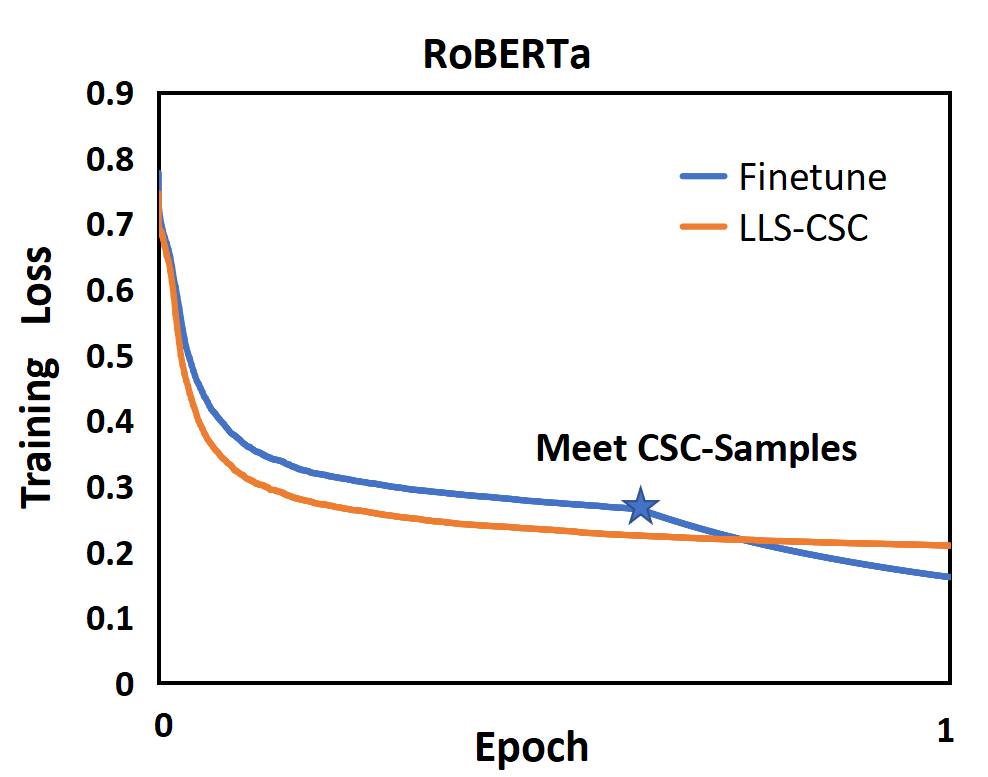}
\label{fig:roberta_loss}
}
\caption{Curves of training loss for ERNIE and RoBERTa. The marker point represents that curve drops rapidly after STM models meet CSC-Samples.}
\label{fig:training_loss_2}
\end{figure}

In Fig.~\ref{fig:training_loss_2}, we show the training loss curve of ERNIE and RoBERTa.

\section{Proportion of CSC-Samples.}\label{app_proportion}
\begin{figure}[t]
\centering
\subfigure{
\includegraphics[scale=0.32]{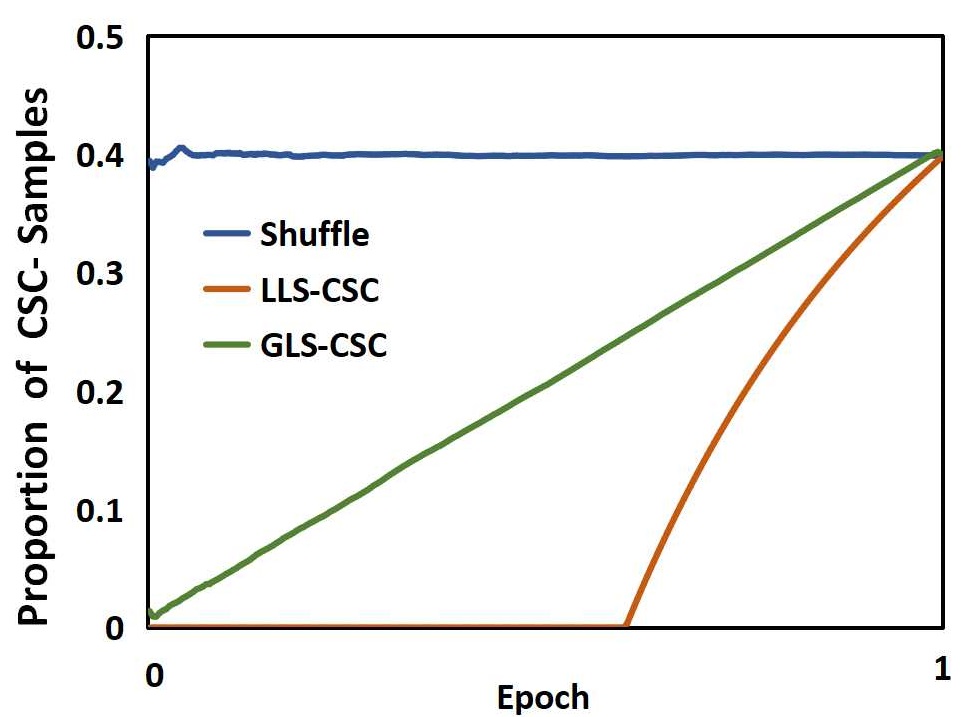}
}
\caption{Proportion of CSC-Samples during the process of training for finetuning, LLS-CSC and GLS-CSC methods.}
\label{fig:proportion_e}
\end{figure}

In Fig.~\ref{fig:proportion_e}, we show the proportion change of CSC-Samples.

\section{Spearman Correlation between Datasets.}\label{app_spearman}
\begin{figure}[t]
\centering
\subfigure[Samples labeled as 0]{
\includegraphics[scale=0.33]{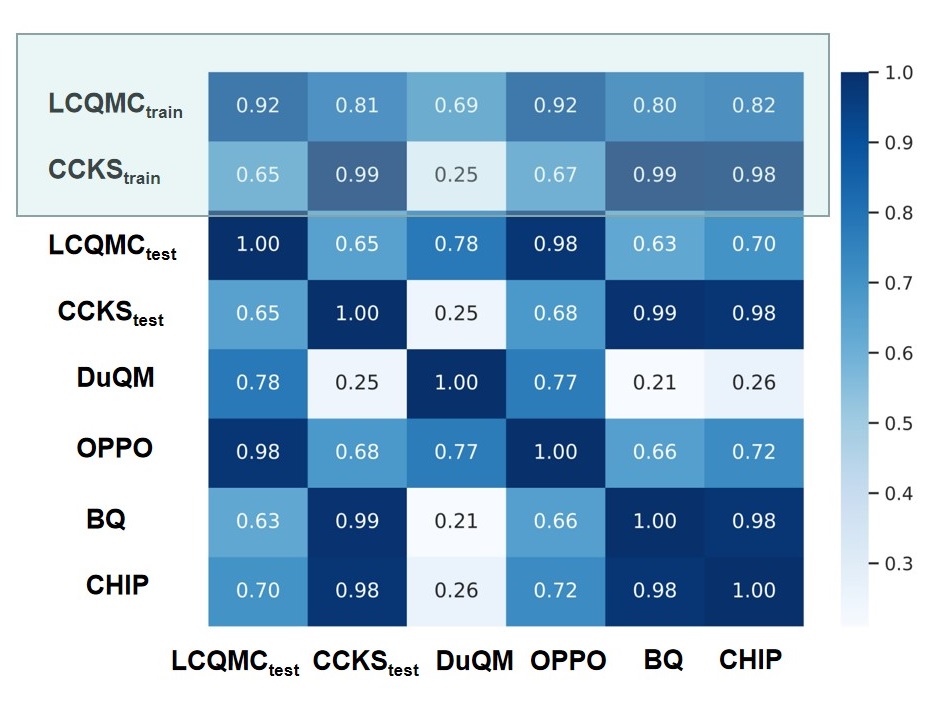}
}
\subfigure[Samples labeled as 1]{
\includegraphics[scale=0.33]{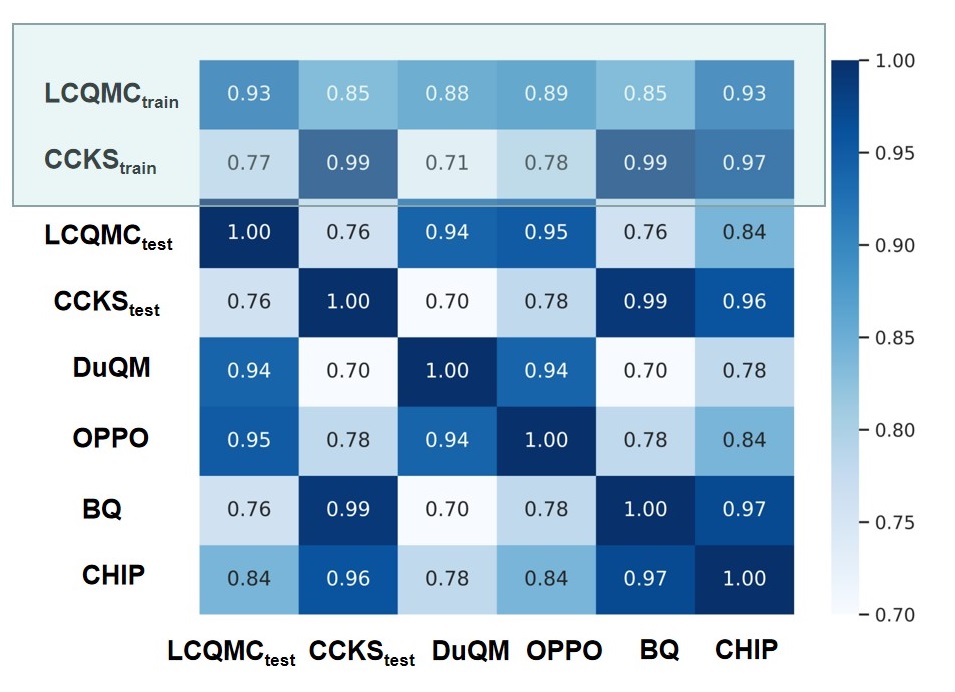}
}
\caption{We count the distribution of edit distance features from samples labeled as 0 and labeled as 1 respectively, and calculate the Spearman's correlation between different datasets with p-value \textless 0.05. The first two rows (covered by shadow) represent the correlation between the training set and the test set, and the rest rows represent the correlation between the different test sets.}
\label{fig:spearman_static}
\end{figure}

We count the distribution of the edit distance feature from samples labeled as 0 and labeled as 1 respectively. In Fig.~\ref{fig:spearman_static}, we calculate the Spearman's correlation between different datasets. We observe significant variance across datasets, so it is valuable for us to conduct a comprehensive O.O.D. evaluation.


\newcommand{\Tabi}[2]{\begin{tabular}{@{}#1@{}}#2\end{tabular}}
\begin{table*}[htb]
\resizebox{\textwidth}{!}
{
\begin{tabular}{ccccc}
\toprule[0.7pt]
                     Type                   & {Text1} & {Text2} & Label & Edit distance \\
\midrule[0.5pt]        
\multirow{3}{*}{Easy-to-predcit sample} & {\Tabi{c}{苹果手机通讯录如何删除 \\ (How to delete iPhone contacts)}}            & {\Tabi{c}{苹果手机电话簿如何删除 \\ (How to delete iPhone phone book)}}            & 1     & 3             \\
\cmidrule[0.5pt]{2-5}
                                        & {\Tabi{c}{属兔的人适合居住在中国哪个城市? \\ (Which city in China is suitable for Rabbit people to live in?)}}       & {\Tabi{c}{中国哪个城市最适合居住？\\ (Which Chinese city is the best to live in?)}}           & 0     & 14            \\
\midrule[0.5pt]
\multirow{3}{*}{Hard-to-predcit sample} & {\Tabi{c}{游戏内无法发送文字消息的原因\\(Reasons why in-game text messages cannot be sent)}}       & {\Tabi{c}{为什么我游戏里面不能发文字呢\\(Why can't I send text in the game?)}}         & 1     & 14            \\
\cmidrule[0.5pt]{2-5}
                                        & {\Tabi{c}{猫喜欢吃什么水果\\(What fruits do cats like to eat)}}              & {\Tabi{c}{牛喜欢吃什么水果\\(What fruits do cows like to eat)}}               & 0     & 1 \\
\bottomrule[0.7pt]
\end{tabular}
}
\caption{Examples for easy-to-predict samples and hard-to-predict samples.}
\label{example}
\end{table*}

\begin{figure}[t]
\centering
\includegraphics[scale=0.32]{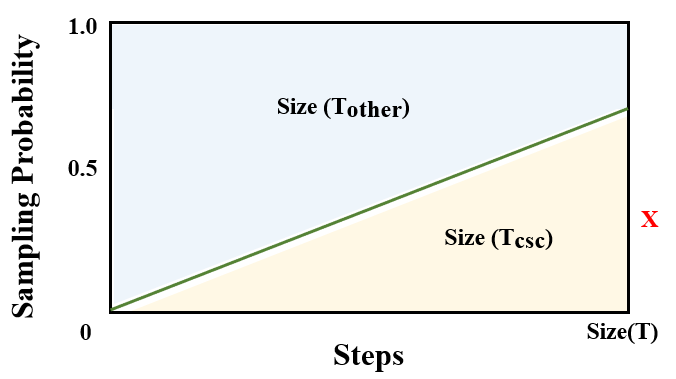}
\caption{For LCQMC$_{train}$, the label distribution of samples at each edit distance.}
\label{fig:tuidao}
\end{figure}

\section{Derivation.}\label{app_derivation}

In Alg.~\ref{algo}, the purpose of $\alpha$ is to control the probability of sampling from the two sets $T_{csc}$ and $T_{other}$ to achieve an ideal situation. Therefore, we derive the initial $\alpha$ value by analyzing the ideal situation.

As shown in Fig.~\ref{fig:tuidao}, it illustrates an ideal situation. The area covered by blue represents the $T_{other}$ set, and covered by yellow represents the $T_{csc}$ set. The horizontal axis represents the number of sampling steps, that is, the size of the training set ($Size(T)$), and the vertical axis represents the probability of sampling. Geometrically, the slope of the green slash represents the value of $\alpha$. Therefore, we can turn the arithmetic problem into a geometric problem. We first set the length of one side of the triangle below as x. Then the area of the triangle and trapezoid can be represented with x respectively. The area of the triangle is
\begin{gather}
\small
Area_{tri}=\frac{x \times Size(T)}{2}
\end{gather}
The area of the trapezoid is
\begin{gather}
\small
Area_{trap}=\frac{(1-x+1) \times Size(T)}{2}
\end{gather}
Then, the ratio of the areas between them can be represented by
\begin{gather}
\small
k=\frac{Area_{tri}}{Area_{trap}} =\frac{(2-x)}{x}
\end{gather}
After conversion, we obtain that x is
\begin{gather}
\small
x=\frac{2}{k+1}
\end{gather}
Above all, the slope of the green slash (the value of $\alpha$) can be represented by tangent angle representation of the triangle, that
\begin{gather}\label{alpha}
\small
\alpha=\frac{x}{Size(T)}=\frac{2}{(k+1) \times Size(T)}
\end{gather}
Furthermore, the areas of the triangle and trapezoid can also be represented as $Size(T_{csc})$ and $Size(T_{other})$, $k$ can also be represented by
\begin{gather}\label{area}
\small
k=\frac{Size(T_{other})}{Size(T_{csc})}
\end{gather}
Finally, bring Eq.~\ref{area} into Eq.~\ref{alpha} to calculate the initial value of $\alpha$.

\begin{table}[t]
\small
\centering
{
\begin{tabular}{l|ccc}
\toprule[0.7pt]
  Dataset    & Train\#  & Dev\#  & Test\#  \\
\midrule[0.5pt]
LCQMC & 238,766 & 8,802 & 12,500 \\
CCKS  & 80,000  & 8,042 & 10,000 \\
DuQM  & -      & -    & 10,121 \\
OPPO  & -      & -    & 10,000 \\
BQ    & -      & -    & 20,000 \\
CHIP  & -      & -    & 4,000 \\
\toprule[0.7pt]
\end{tabular}
}
\caption{Statistics of datasets used in our experiments.}
\label{app:statistic}
\end{table}

\end{CJK*}
\end{document}